\documentclass{article}
\usepackage{ijcai952e,dialog2e,algorithm,epsf,named,times}

\title{Generating Information-Sharing Subdialogues in Expert-User
Consultation\thanks{This material is based upon work supported by the
National Science Foundation under Grant No. IRI-9122026.}}

\author{Jennifer Chu-Carroll \and Sandra Carberry \\
        Department of Computer and Information Sciences \\
        University of Delaware \\
        Newark, DE 19716, USA \\
        E-mail: \{jchu,carberry\}@cis.udel.edu}

\date{}

\begin{document}

\maketitle

\begin{abstract}

In expert-consultation dialogues, it is inevitable that an agent will
at times have insufficient information to determine whether to accept
or reject a proposal by the other agent. This results in the need for
the agent to initiate an information-sharing subdialogue to form a set
of shared beliefs within which the agents can effectively re-evaluate
the proposal. This paper presents a computational strategy for
initiating such information-sharing subdialogues to resolve the
system's uncertainty regarding the acceptance of a user proposal. Our
model determines when information-sharing should be pursued, selects a
focus of information-sharing among multiple uncertain beliefs, chooses
the most effective information-sharing strategy, and utilizes the
newly obtained information to re-evaluate the user
proposal. Furthermore, our model is capable of handling embedded
information-sharing subdialogues.

\end{abstract}

\section{Introduction}
\label{intro}

We have been studying a particular kind of collaborative dialogue in
which two participants (a consultant and an executing agent)
collaborate on developing a plan to achieve the executing agent's
domain goal.  In such an environment, the consultant and the executing
agent have different knowledge about the domain and about the
executing agent's particular circumstances and preferences that may
affect the domain plan being constructed. Thus, it is inevitable that
an agent will not always immediately accept the actions or beliefs
proposed by the other agent. However, an agent should recognize the
collaborative nature of the interaction and the fact that each agent
has private knowledge that is not shared by the other agent. Thus,
rather than indiscriminately rejecting proposals that she does not
have sufficient reasons to accept, a collaborative agent should both
share her private knowledge with the other agent and solicit relevant
information from the other agent in order for both agents to
effectively re-evaluate the proposal and come to the most beneficial
decision.

Such collaborative information-sharing behaviour is illustrated in the
following dialogue segment based on transcripts of naturally occuring
dialogues \cite{sri92}. In this dialogue, a travel agent (T) and a
customer (C) are constructing a plan for two other agents to travel
from San Francisco to Los Angeles. This segment follows a proposal
that the travelers be booked on a particular USAir flight.

\bdialog{T:}{C:} \em

\speakerlab \label{american} Can we put them on American?

\listenerlab \label{why} Why?

\speakerlab We're having a lot of problems on the USAir seat maps so I
don't know if I can get them together.
\dialine But American whatever we request pretty much we get.

\listenerlab \label{care} I don't know if they care if they sit
together. 
\dialine \label{usair} Let's go ahead and stick with USAir.

\edialog In this dialogue, T proposes putting the travelers on
American Airlines instead of USAir in utterance (\ref{american}). In
(\ref{why}), C questions T's motivation for this proposed action ---
i.e., the support that T's private knowledge provides for this
proposal. After T provides her motivation, C informs T in (\ref{care})
that she rejects the motivation, re-evaluates the proposal, and in
(\ref{usair}) rejects the actions proposed by T.

This paper presents a computational model for collaborative
information-sharing during proposal evaluation. Our model first uses
the system's existing beliefs along with evidence provided by the user
to evaluate user proposals and to determine whether they should be
accepted or rejected. If the system has insufficient information to
make this decision, it initiates an information-sharing subdialogue to
form a set of shared beliefs within which the agents can effectively
re-evaluate the proposal and come to agreement. This may lead to
evaluation of an agent's reasons for a proposal and further
information-sharing about an agent's beliefs supporting these reasons,
thus leading to an embedded information-sharing subdialogue. 

Our research contributes to response generation in collaborative
interaction by 1) providing an algorithm for identifying when an
information-sharing subdialogue should be initiated during proposal
evaluation, 2) providing a selection algorithm for determining the
beliefs that should be the focus of information-sharing, 3)
formulating information-sharing strategies and identifying the
criteria for invoking each strategy, and 4) capturing the process in a
{\em Propose-Evaluate-Modify} cycle that enables embedded
information-sharing subdialogues.

\section{Modeling Collaborative Activities}

In modeling collaborative activities, it is essential that the system
captures the agents' intentions conveyed by their utterances.  Our
model utilizes an enhanced version of the dialogue model described in
\cite{lam_car_acl91} to represent the current status of the
interaction. The enhanced dialogue model has four levels: the {\em
domain} level which consists of the domain plan being constructed for
later execution, the {\em problem-solving} level which contains the
actions being performed to construct the domain plan, the {\em belief}
level which consists of the mutual beliefs pursued to further the
problem-solving intentions, and the {\em discourse} level which
contains the communicative actions initiated to achieve the mutual
beliefs \cite{chu_car_aaai94}.

In our earlier work, we developed a plan-based model that captures
collaborative planning in a {\em Propose-Evaluate-Modify} cycle of
actions \cite{chu_car_aaai94}. This model treats a collaborative
planning process as a sequence of the following actions: agent A's
{\em proposal} of a set of actions and beliefs to be added to the {\em
shared plan} \cite{gro_sid_ic90,all_snlw91} being developed, agent B's
{\em evaluation} of the proposed actions and beliefs, and B's proposed
{\em modifications} to the original proposal in cases where the
proposal is rejected. Notice that B's proposed modifications will
again be evaluated by A, and if conflicts arise, A may propose
modifications to B's proposed modifications, resulting in a recursive
process.

However, our previous research assumed that an agent's evaluation of a
proposal always results in the proposal being accepted or rejected,
and did not take into account cases in which the agent initially has
insufficient information to determine whether or not to accept the
proposal, as shown in the example in utterances
(\ref{american})-(\ref{usair}).  This paper extends our earlier work
by providing a computational strategy for collaborative
information-sharing during proposal evaluation. We focus on situations
in which the system's lack of knowledge occurs during the evaluation
of proposals at the belief level of the dialogue model.\footnote{We
are concerned with situations in which the system recognizes the
user's proposal but cannot decide whether to accept or reject it, not
those where the system initiates a clarification subdialogue to
disambiguate the user's proposal
\cite{vbetal_ci93,logetal_tr94,hee_hir_tr92,ras_zuk_cogsci93}.}

\section{Information-Sharing During Collaboration}

Since a collaborative agent initiates information-sharing subdialogues
to help determine whether to accept or reject a proposed belief, the
information-sharing process is captured as part of the {\em
evaluation} process in the {\em Propose-Evaluate-Modify} cycle for
collaborative activities. Thus the evaluation of a proposed belief
involves the agent 1) determining the acceptance of the proposed
belief based on the information currently available to her, and 2) in
cases where she cannot decide whether to accept or reject the belief,
initiating an information-sharing subdialogue so that the agents can
exchange information and re-evaluate the proposed belief. The
following sections describe these two processes.

\subsection{Evaluating Proposed Beliefs}
\label{evaluate}

Our system maintains a set of beliefs about the domain and about the
user's beliefs. Associated with each belief is a {\em strength} that
represents the agent's confidence in holding the belief. We model the
strength of a belief using {\em endorsements} \cite{coh_book85},
following \cite{gal_br92,logetal_tr94}, based on the semantic form of
the utterance used to convey a belief, the level of expertise of the
agent conveying the belief, stereotypical beliefs, etc.

\begin{figure}
\footnotesize

{\bf Evaluate-Belief(\_bel):}
\begin{algorithm}

\item evidence set $\leftarrow$ \_bel (appropriately endorsed as
conveyed by the user) and the system's beliefs that support or attack
\_bel.

\item If \_bel has no children, return {\bf Evaluate}(\_bel, evidence
set). 

\item Evaluate each of \_bel's children: \_bel$_1$, \ldots, \_bel$_n$:

   \begin{algorithm}

   \item belief\_result $\leftarrow$ {\bf Evaluate-Belief}(\_bel$_i$)

   \item rel\_result $\leftarrow$ {\bf
Evaluate-Belief}(supports(\_bel$_i$,\_bel))  

   \item If belief\_result = reject or rel\_result = reject, ignore
\_bel$_i$ and supports(\_bel$_i$,\_bel).  

   \item Else if belief\_result = rel\_result = accept, add
\{\_bel$_i$, supports(\_bel$_i$,\_bel)\} to the evidence set.

   \item Else if belief\_result = unsure or rel\_result = unsure, add
\{\_bel$_i$, supports(\_bel$_i$,\_bel)\} to the potential evidence
set.

   \end{algorithm}

\item Evaluate \_bel:

   \begin{algorithm}

   \item upperbound $\leftarrow$ {\bf Evaluate}(\_bel, evidence set +
potential evidence set)

   \item lowerbound $\leftarrow$ {\bf Evaluate}(\_bel, evidence set) 

   \item If upperbound = lowerbound = accept, accept \_bel.
   
   \item Else if upperbound = lowerbound = reject, reject \_bel.

   \item Else unsure about \_bel; annotate \_bel with upperbound,
lowerbound, evidence set, and potential evidence set.

   \end{algorithm}

\end{algorithm}
\caption{Algorithm for Evaluating a Belief}
\label{evaluate_algo}
\vspace{1ex}
\hrule
\end{figure}

The belief level of our dialogue model consists of one or more belief
trees where the belief represented by a child node is intended to
support that represented by its parent. When an agent proposes a new
belief and gives (optional) supporting evidence for it, this set of
proposed beliefs is represented as a belief tree. The system must then
evaluate the proposed beliefs in order to determine whether to accept
the proposal, reject it, or pursue information-sharing to allow the
agents to re-evaluate it. The algorithm for evaluating proposed
beliefs is shown in Figure~\ref{evaluate_algo}, and is applied to the
root node of each proposed belief tree (the top-level proposed
beliefs). Since the acceptance of a child belief may affect the
acceptance of its parent, before determining the acceptance of a
belief or evidential relationship, its children in the proposed belief
tree must be evaluated (step 3). Thus, for each child belief of \_bel,
the system evaluates both the belief (step 3.1) and the evidential
relationship between the belief and \_bel (step 3.2). A piece of
evidence is marked as 1) accepted if both the child belief and the
evidential relationship are accepted, 2) rejected if either the child
belief or the evidential relationship is rejected, and 3) uncertain
otherwise.

To determine the status (accepted, rejected, or uncertain) of a belief
\_bel, the algorithm constructs an {\em evidence set} that contains
the user's proposal of \_bel, endorsed according to the user's level
of expertise in that subarea as well as the user's strength in the
belief as conveyed by the semantic form of the utterance (step 1), the
system's own beliefs pertaining to \_bel (step 1), and evidence
proposed by the user that is accepted by the system (step 3.4). It
also constructs a {\em potential evidence set} consisting of evidence
proposed by the user whose acceptance is undetermined (step 3.5). The
algorithm must then determine whether the potential evidence could
have an impact on the system's decision-making. It first evaluates
\_bel by invoking the {\bf Evaluate} function\footnote{{\bf Evaluate}
utilizes a simplified version of Galliers' belief revision mechanism
\cite{gal_br92,logetal_tr94} which, given a set of evidence, compares
the endorsements of the beliefs that support and attack \_bel and
determines whether or not \_bel should be accepted.}  to compute an
upperbound and a lowerbound for the system's acceptance of \_bel. The
upperbound is computed by invoking the {\bf Evaluate} function with
evidence from both the evidence set and the potential evidence set,
i.e., treating all uncertain evidence as accepted, and the lowerbound
is computed by invoking {\bf Evaluate} with only the evidence set,
i.e., treating all uncertain evidence as rejected (steps 4.1 and
4.2). If \_bel is either accepted or rejected in both cases,
indicating that the uncertainty of the evidence, if any, does not
affect the acceptance of \_bel, the system accepts or rejects \_bel
(steps 4.3 and 4.4).  Otherwise, the system has insufficient
information to determine the acceptance of \_bel and it is marked as
uncertain (step 4.5). If the top-level proposed belief is marked as
uncertain, an information-sharing subdialogue will be initiated, as
described in the next section.

\subsection{Initiating Information-Sharing Subdialogues}

A collaborative agent, when facing a situation in which she is
uncertain about whether to accept a proposal, should attempt to share
information with the other agent so that each agent can knowledgably
re-evaluate the proposal and the agents can come to agreement --- to
do otherwise is to fail in her responsibilities as a collaborative
agent. Furthermore, a collaborative agent should engage in effective
and efficient dialogues; thus she should pursue the
information-sharing subdialogue that she believes will most likely
result in the agents coming to an intelligent decision about the
proposal. The process for initiating information-sharing subdialogues
involves two steps: selecting a focus of information-sharing from the
proposed beliefs marked as uncertain during the initial evaluation
process, and selecting an effective information-sharing strategy.

\subsubsection{Selecting the Focus of Information-Sharing}

The possible combinations of the upperbound and lowerbound values
produced by the {\bf Evaluate-Belief} algorithm
(Figure~\ref{evaluate_algo}) are shown in
Figure~\ref{combination}.\footnote{Our model assumes that a child
belief is always intended to provide support for its parent belief (a
piece of counterevidence is represented as a child belief supporting
the negation of the parent belief); thus only six out of the nine
theoretically possible combinations may occur.} Cases 1 and 6
correspond to steps 4.3 and 4.4 in Figure~\ref{evaluate_algo},
respectively, in which the decision to accept or reject is the same
whether or not beliefs in the potential evidence set are accepted. In
these cases, the uncertainty in the child beliefs need not be resolved
since their acceptance will not impact acceptance of the parent belief
that they are intended to support, and thus will not affect acceptance
of the top-level proposed belief that is important to the plan being
constructed.\footnote{Young et al. \shortcite{youetal_cogsci94} argued
that if a belief is accepted even though a child belief that is
intended to support it is rejected, the rejection of the child belief
need not be addressed since it is no longer relevant. Our strategy
extends this concept to uncertain information.} In case 4, the system
will remain unsure whether to accept or reject \_bel regardless of
whether the uncertain child beliefs, if any, are accepted or rejected,
i.e., resolving the uncertainty in the child beliefs will not help
resolve the uncertainty in \_bel. Thus, the system should focus on
sharing information to resolve the uncertainty about \_bel itself
instead of its children. In cases 2 and 3, acceptance of the child
beliefs has the potential to influence acceptance of \_bel and in
cases 3 and 5, rejection of the child beliefs can lead to rejection of
\_bel. Thus in all three cases, the system should initiate
information-sharing that will allow the agents to come to agreement
about the currently uncertain child beliefs.

\begin{figure}
\footnotesize
\begin{center}
\begin{tabular}{lccl}
& {\bf Upper} & {\bf Lower} & {\bf Action} \\
1 & accept & accept & accept \_bel \\
2 & accept & unsure & attempt to accept uncertain children in \\
  &&& order to accept \_bel \\ 
3 & accept & reject & actions in cases 2 and 5\\
4 & unsure & unsure & resolve uncertainty regarding \_bel itself \\
5 & unsure & reject & attempt to reject uncertain children in \\
  &&& order to reject \_bel \\  
6 & reject & reject & reject \_bel
\end{tabular}
\end{center}
\caption{Combinations of Upperbounds and Lowerbounds}
\label{combination}
\vspace{1ex}
\hrule
\end{figure}

\begin{figure}
\footnotesize

{\bf Select-Focus-Info-Sharing}(\_bel):

\newcommand{\var}[1]{\mbox{\_#1}}
\begin{algorithm}

\item If \_bel is accepted or rejected, focus $\leftarrow$ \{\},
return focus.

\item If \_bel has no children, focus $\leftarrow$ \_bel; return focus.

\item \label{both_uncertain} If upper = lower = uncertain, focus
$\leftarrow$ \_bel; return focus.

\item \label{upper} If upper = accept,

   \begin{algorithm}

   \item \label{rank} Assign each piece of uncertain evidence
(\_bel$_i$ and supports(\_bel$_i$,\_bel)) to a set, and order the sets
according to how close the beliefs in each set were to being
accepted. Call them \_set$_1$, \ldots, \_set$_m$.

   \item \label{predict} For each set in ranked order, do until
new\_result$_i$ = accept:

   new\_result$_i$ $\leftarrow$ {\bf Evaluate}(\_bel, evidence set + set$_i$)

   \item \label{fail} If new\_result$_i$ $\neq$ accept, increase set
size by one, re-rank sets and goto 4.2;

   \item \label{succeed} Else

         focus $\leftarrow \bigcup_{\var{el}_j \in \var{set}_i}
\mbox{\bf Select-Focus-Info-Sharing}(\var{el}_j)$; return focus.

   \end{algorithm}

\item \label{lower} If lower = reject,

   \begin{algorithm}

   \item Assign each piece of uncertain evidence (\_bel$_i$ and
supports(\_bel$_i$,\_bel)) to a set, and order the sets according to
how close the beliefs in each set were to being rejected. Call them
\_set$_1$, \ldots, \_set$_m$.

   \item For each set in ranked order, do until new\_result$_i$ =
reject: 

   new\_result$_i$ $\leftarrow$ {\bf Evaluate}(\_bel, evidence set +
potential evidence set - set$_i$)

   \item If new\_result$_i$ $\neq$ reject, increase set size by
one, re-rank sets and goto 5.2;

   \item Else

         focus $\leftarrow \bigcup_{\var{el}_j \in \var{set}_i}
\mbox{\bf Select-Focus-Info-Sharing}(\var{el}_j)$; return focus.

   \end{algorithm}

\end{algorithm}
\caption{Selecting the Focus of Information-Sharing}
\label{focus_algo}
\vspace{1ex}
\hrule
\end{figure}

\looseness=-1000 However, there may be more than one uncertain child
belief. Thus when the system initiates information-sharing, it must
first select a belief on which to focus during the information-sharing
process. Our algorithm for selecting the focus of information-sharing
is shown in Figure~\ref{focus_algo}. {\bf Select-Focus-Info-Sharing}
is initially invoked with \_bel instantiated as the top-level proposed
belief. Step~\ref{both_uncertain} of the algorithm corresponds to case
4 in Figure~\ref{combination} where the uncertainty in the child
beliefs is irrelevant to the acceptance of \_bel; thus the focus of
information-sharing is \_bel itself. Steps~\ref{upper} and \ref{lower}
of the algorithm correspond to cases 2, 3 and 5 in
Figure~\ref{combination} where the system attempts to share
information to resolve the uncertainty in the child beliefs and
perhaps thereby accept or reject \_bel.

Step~\ref{upper} of the algorithm is concerned with cases where the
potential acceptance of uncertain child beliefs may lead to the
acceptance of \_bel (cases 2 and 3 in Figure~\ref{combination}). In
selecting the focus of information-sharing in such cases, two factors
should come into play: 1) how strongly the acceptance of each piece of
evidence affects the acceptance of \_bel --- the stronger the impact
that the potential evidence can have on the acceptance of \_bel, the
more useful it is to expend effort on resolving the uncertainty about
the proposed evidence, and 2) how close each piece of evidence was
to being accepted during the initial evaluation process --- the closer
a piece of evidence is to being accepted, the easier it is for the
system to gather sufficient information to accept the evidence. Our
algorithm first constructs a singleton set for each piece of uncertain
evidence for \_bel, where a piece of evidence includes a pair of
beliefs: a child belief,$ \_bel_i$, and the evidential relationship
between \_bel$_i$ and \_bel, {\em supports(\_bel$_i$,\_bel)}. The sets
are ordered according to how close the beliefs in a set were to being
accepted in the initial evaluation process (step~\ref{rank}). The
first set (\_set$_1$) is then added to the evidence set, and \_bel is
re-evaluated with respect to the augmented evidence set
(step~\ref{predict}), thus considering the potential effect of the
acceptance of beliefs in \_set$_1$ on the acceptance of \_bel. If the
result of the evaluation is to accept \_bel, indicating that resolving
the uncertainty of the beliefs in \_set$_1$ is sufficient to resolve
the uncertainty of \_bel, then {\bf Select-Focus-Info-Sharing} is
recursively applied to each belief in \_set$_1$ in order to determine
the focus for resolving the uncertainty of beliefs in \_set$_1$
(step~\ref{succeed}). On the other hand, if the evaluation indicates
that accepting \_set$_1$ does not result in the acceptance of \_bel,
the next set (\_set$_2$) is tried. This continues until either the
uncertain evidence in a set is predicted to resolve the uncertainty of
\_bel, or all of the uncertain evidence is tried and none suffices for
acceptance of \_bel. In the latter case, the set size is increased by
one, sets of the requisite size are constructed by combining
individual pieces of evidence, the new sets are ordered, and the same
process is repeated (step~\ref{fail}). Thus our algorithm guarantees
that the fewest possible beliefs are selected as the focus of
information-sharing and that these beliefs require the least effort to
achieve among those that are strong enough to affect the acceptance of
\_bel.\footnote{In cases where the focus set contains multiple
beliefs, additional processing is needed to determine the most
coherent order in which to address the beliefs.}

Step~\ref{lower} of the algorithm corresponds to cases 3 and 5 in
Figure~\ref{combination}. The procedure for step~\ref{lower} is
similar to that for step~\ref{upper} except that in predicting the
effect of resolving a piece of uncertain evidence, \_bel is evaluated
under the assumption that the set of uncertain evidence under
consideration is rejected while the other uncertain beliefs are
accepted (step 5.2).

\subsubsection{Selecting an Information-Sharing Strategy}

We have identified four strategies which a collaborative agent may
adopt in initiating an information-sharing subdialogue to allow the
agents to share information and re-evaluate a belief or evidential
relationship, \_bel:

\label{strategies}
\begin{enumerate}

\item Agent A may present a piece of evidence against \_bel and
(implicitly) invite agent B to attack it. Such a strategy focuses B's
attention on the counterevidence and suggests that it is what keeps A
from accepting \_bel. Thus in collaborative activities, this strategy
should only be employed if A's counterevidence is {\em critical}, i.e.,
if proving that the counterevidence is invalid will cause A to accept
\_bel. This strategy also allows the possibility of B accepting the
counterevidence and perhaps both agents subsequently adopting
$\lnot$\_bel instead of \_bel.

\item Agent A may query B about his reasons for believing in
\_bel. This strategy is appropriate when A does not know B's support
for \_bel, and also does not have evidence against \_bel herself. It
would result either in A gathering evidence that contributes toward
her adopting \_bel, or in A discovering B's invalid justification for
holding \_bel and attempting to convince B of $\lnot$\_bel.

\item Agent A may query B for his evidence for \_bel and also present
her reasons for believing in $\lnot$\_bel. This strategy is adopted
when A does not know B's reasons for believing \_bel, but does have
non-critical evidence against accepting \_bel. In this case B may
provide his support for \_bel, attack A's evidence against \_bel, or
accept A's counterevidence and perhaps subsequently adopt
$\lnot$\_bel.

\item Agent A may indicate her uncertainty about \_bel and present her
reasons against \_bel. This strategy is adopted when A is least
certain about how to go about sharing information to resolve the
uncertainty --- when A already knows B's reasons for believing \_bel,
and only has non-critical evidence against accepting \_bel. In a
collaborative environment, A's indication of the uncertainty in her
decision should lead B to provide information that he believes will
help A re-evaluate the proposal.

\end{enumerate}

The process for initiating information-sharing subdialogues is
performed by invoking the {\em Share-Info-Reevaluate-Belief}
problem-solving action on the focus identified by {\bf
Select-Focus-Info-Sharing} (Figure~\ref{focus_algo}). It initiates an
information-sharing subdialogue using the most appropriate of the four
information-sharing strategies and re-evaluates the top-level belief
taking into account the newly obtained information. The
recipe\footnote{A recipe \cite{pol_acl86} is a template for performing
actions. It contains the applicability conditions for performing an
action, the subactions comprising the body of an action, etc.} for
{\em Share-Info-Reevaluate-Belief} specifies that in order for the
action to be invoked, it must be the case that the system believes in
neither a top-level proposed belief (\_bel) nor its negation --- that
is, the system cannot determine whether to accept or reject \_bel. The
body of {\em Share-Info-Reevaluate-Belief} consists of alternative
subactions which correspond to the aforementioned strategies that a
collaborative agent can use to pursue information-sharing. The recipes
for two of these subactions are shown in Figure~\ref{recipes}.

\begin{figure}
\footnotesize
\begin{tabbing}
Appl Cond: \= xx \= Reevaluate-After-Invite-Attack( \= \kill
Action: \>
Reevaluate-After-Invite-Attack(\_s1,\_s2,\_bel1,\_top-belief,\\
        \> \> \> \_belief-tree) \\ 
Type: \> Decomposition \\
Appl Cond: \> uncertain(\_s1,\_bel1)\\
           \> believe(\_s1,\_bel2) \\
           \> believe(\_s1,supports(\_bel2,$\lnot$\_bel1)) \\
           \> results-in(believe(\_s1,$\lnot$\_bel2),believe(\_s1,\_bel1)) \\
Constraint: \> member-of(\_bel1,\_belief-tree)\\
            \> root-of(\_belief-tree,\_top-belief)\\
Preconds: \> MB(\_s1,\_s2,\_bel2) $\land$
MB(\_s1,\_s2,supports(\_bel2,$\lnot$\_bel1))\\
        \> \>  $\lor$ MB(\_s1,\_s2,$\lnot$\_bel2) \\
        \> \> $\lor$ MB(\_s1,\_s2,$\lnot$supports(\_bel2,$\lnot$\_bel1))\\
Body: \> Evaluate-Proposed-Beliefs(\_s1,\_s2,\_top-belief) \\
Goals: \> believe(\_s1,\_top-belief) $\lor$ believe(\_s1,$\lnot$\_top-belief)
\\ 

\\

Action: \> Reevaluate-After-Ask-Why(\_s1,\_s2,\_bel1,\_top-belief,\\
        \> \> \> \_belief-tree) \\
Type: \> Decomposition \\
Appl Cond: 
\>
$\lnot$knowref(\_s1,\_bel2, believe(\_s2,supports(\_bel2,\_bel1))\\
\> $\lnot$knowref(\_s1,\_bel3, supports(\_bel3,$\lnot$\_bel1))
\\
Constraint: \> member-of(\_bel1,\_belief-tree)\\
            \> root-of(\_belief-tree,\_top-belief)\\
        
Preconds:
\> knowref(\_s1,\_bel2, believe(\_s2,\_supports(\_bel2,\_bel1)))\\
Body: \> Evaluate-Proposed-Beliefs(\_s1,\_s2,\_top-belief) \\
Goals: \> believe(\_s1,\_top-belief) $\lor$ believe(\_s1,$\lnot$\_top-belief)
\end{tabbing}

\caption{Specializations of {\em Share-Info-Reevaluate-Belief}}
\label{recipes}
\vspace{1ex}
\hrule
\end{figure}

The first specialization, {\em Reevaluate-After-Invite-Attack},
corresponds to the first information-sharing strategy in which the
system (\_s1) has a piece of {\em critical evidence} (\_bel2) against
believing \_bel1, a belief proposed by the user (\_s2) and about which
the system is uncertain. This criterion is captured in the
applicability conditions\footnote{Applicability conditions are
conditions that must already be satisfied in order for an action to be
reasonable to pursue, whereas an agent can try to achieve unsatisfied
preconditions.} of the action: the conditions that the system is
uncertain about the acceptance of \_bel1, that the system believes in
\_bel2 which provides support for $\lnot$\_bel1, and that the system's
disbelief in \_bel2 will result in its adoption of \_bel1. The
preconditions of {\em Reevaluate-After-Invite-Attack}, however, show
that the action cannot be performed until one of the following
conditions is true: 1) the system and the user mutually believe (MB)
in \_bel2 and mutually believe that \_bel2 supports $\lnot$\_bel1, 2)
the system and the user mutually believe in $\lnot$\_bel2, or 3) the
system and the user mutually believe that \_bel2 does not support
$\lnot$\_bel1. In order to satisfy the preconditions, the system will
adopt the {\em Express-Doubt} discourse action \cite{lam_car_acl92},
in which the system expresses doubt at \_bel1 by contending \_bel2 and
the evidential relationship between \_bel2 and $\lnot$\_bel1, as an
attempt to achieve {\em MB(S,U,\_bel2)} and {\em MB(S,U,
supports(\_bel2,$\lnot$\_bel1))}.\footnote{These two mutual beliefs
are selected as preconditions to be satisfied because the system
itself holds these beliefs. The alternative preconditions are present
in order to capture situations in which the user, in response to the
{\em Express-Doubt} action, convinces the system that \_bel2 is false
or that \_bel2 does not serve as justification for $\lnot$\_bel1. The
use of these alternative preconditions will be demonstrated in
Section~\ref{possible}.} Thus the system will initiate an
information-sharing subdialogue by expressing its evidence against the
proposed belief and inviting the user to comment on it. If the outcome
of the information-sharing subdialogue satisfies one of the
preconditions of {\em Reevaluate-After-Invite-Attack}, the system can
perform the body of the action and re-evaluate \_top-belief (the root
node of the proposed belief tree of which \_bel1 is a part) taking
into account the newly obtained information. Notice that the user's
response to the {\em Express-Doubt} discourse action is again
considered a proposal of mutual beliefs and will be evaluated by the
system. The system may again have insufficient information to
determine whether to accept or reject the new proposal which was
intended to resolve the uncertainty of the previous proposal. It will
then initiate another information-sharing subdialogue to resolve the
new uncertainty, resulting in embedded information-sharing
subdialogues.

\looseness=-1000
The second specialization of {\em Share-Info-Reevaluate-Belief} is
{\em Reevaluate-After-Ask-Why}, which corresponds to the second
information-sharing strategy in which the system attempts to find out
the user's justification for believing \_bel1. The action is
applicable (Figure~\ref{recipes}) if the system (\_s1) does not know
the user's (\_s2's) justification for holding \_bel1, and also does
not have any evidence against \_bel1 itself. We argue that a
collaborative agent should not accept a proposed belief merely because
of the lack of evidence to the contrary. Instead, she should only
accept a belief if the evidence supporting the belief is strong enough
to warrant acceptance. For instance, suppose a student informs his
advisor that the AI course scheduled for next semester has been
canceled, without giving any justification for it (such as attributing
the source of the knowledge); although the advisor may not have
evidence against believing in the cancellation, she does not
immediately accept the proposed belief because, given the student's
presumed low expertise in the domain, the endorsement attached to the
proposed belief is not reliable enough to warrant acceptance. The
precondition of {\em Reevaluate-After-Ask-Why} indicates that the
action can be performed only if the system knows the user's evidence
for holding \_bel1. In order to satisfy this precondition, the system
will adopt discourse actions to query the user for such information,
thus initiating an information-sharing subdialogue.

\section{Example}
\label{examples}

Suppose that the system, an expert in the university course advisement
domain, has proposed the options of taking Logic or Algorithms to
satisfy the user's core course requirement. Consider the following
continuation which illustrates many of the features of our strategy
for information-sharing during proposal evaluation:

\bdialogcont{U:}{S:}
\em

\speakerlab \label{better} Logic is a better choice than Algorithms.

\dialine \label{smith} Dr.~Smith is teaching Logic.

\listenerlab \label{sabbatical} Isn't Dr.~Smith going on sabbatical
next year?

\speakerlab \label{postpone} I thought he postponed his sabbatical
until 1996.

\listenerlab \label{why2} Why do you think Dr.~Smith postponed his
sabbatical until 1996?

\dialine \label{ibm} Isn't he spending next year at IBM?

\edialog

\noindent In utterance (\ref{sabbatical}), S initiates an
information-sharing subdialogue to determine whether to accept or
reject the belief that Dr.~Smith is teaching Logic, proposed by U in
(\ref{smith}), by expressing a strong but not warranted
belief\footnote{The strength of a belief falls into one of three
categories: {\em warranted}, {\em strong}, or {\em weak}, based on the
endorsements of the belief.} that Dr.~Smith is going on sabbatical
next year. In (\ref{postpone}), U initiates an information-sharing
subdialogue to determine whether to accept S's claim that Dr.~Smith is
going on sabbatical next year by expressing his weak belief that
Dr.~Smith has postponed his sabbatical. Finally, in (\ref{why2}) and
(\ref{ibm}), S initiates an information-sharing subdialogue to
determine whether to accept U's claim that Dr.~Smith has postponed his
sabbatical by explicitly querying U's reasons for holding this belief
and expressing her belief that Dr.~Smith is spending next
year at IBM. The following sections describe how our model will
produce these information-sharing subdialogues.

\subsection{Evaluating Utterances (\protect{\ref{better}}) and
(\protect{\ref{smith}})} 

Utterances (\ref{better}) and (\ref{smith}) propose two mutual
beliefs, {\em Better-Than(Logic, Algorithms)} and {\em
Teaches(Smith,Logic)}, as well as an evidential relationship that the
latter provides support for the former. When presented these proposed
beliefs, the system will first determine whether to accept or reject
the proposal by invoking {\bf Evaluate-Belief}
(Figure~\ref{evaluate_algo}) on the top-level proposed belief, {\em
Better-Than(Logic,Algorithms)}. The system will evaluate the proposed
evidence as part of evaluating the belief (step 3 in
Figure~\ref{evaluate_algo}), thus recursively invoking {\bf
Evaluate-Belief} on {\em Teaches(Smith,Logic)} (step 3.1) and the
proposed evidential relationship (step 3.2). Since {\em
Teaches(Smith,Logic)} has no children in the proposed belief tree, it
will be evaluated by a simplified version of Galliers' belief revision
mechanism \cite{gal_br92} (step 2). Suppose that the system has the
following evidence pertaining to {\em Teaches(Smith,Logic)}: 1) a
strong belief that Dr. Smith usually teaches Logic, 2) a strong belief
that Dr. Smith is going on sabbatical next year and a warranted belief
that going on sabbatical implies that a faculty member is not teaching
courses, and 3) the user's belief that Dr. Smith is teaching
Logic. The strengths of evidence for and against {\em
Teaches(Smith,Logic)} will be combined and compared. In this case, the
strengths of the two sets of evidence are relatively comparable; thus
the system will not be able to decide whether to accept or reject {\em
Teaches(Smith,Logic)} based on the available information. The system
will then evaluate the proposed evidential relationship (step
3.2). Since the system believes that 1) the user believes that
Dr. Smith is a good teacher and 2) students generally prefer courses
taught by good teachers, Dr. Smith teaching Logic provides support for
the user preferring Logic to Algorithms; thus the proposed evidential
relationship will be accepted. Since the proposed evidential
relationship is accepted while the child belief is uncertain, this
piece of evidence will be added to the potential evidence set (step
3.5).

\looseness=-1000
The system will then evaluate the top-level proposed belief, taking
into account the result of evaluating its only piece of evidence
provided by the user. The system's evidence set for {\em
Better-Than(Logic, Algorithms)} consists of a warranted belief that
Algorithms is a pre-requisite for more CS courses than Logic is, which
provides some support for Algorithms being a better choice than Logic,
as well as the user's statement that Logic is a better choice than
Algorithms. The potential evidence set consists of a pair of beliefs:
the uncertain belief that Dr. Smith is teaching Logic and the accepted
evidential relationship that Dr. Smith teaching Logic provides support
for Logic being a better choice than Algorithms. When both the
evidence set and the potential evidence set are included in the
evaluation, the system will compute the upperbound of the acceptance
of {\em Better-Than(Logic,Algorithms)} to be accept. When considering
only evidence from the evidence set, however, the system will be
uncertain about the acceptance of the proposed belief. The result of
this evaluation corresponds to case 2 in Figure~\ref{combination}, and
results in the need for the system to initiate an information-sharing
subdialogue to resolve the uncertainty.

Since the system cannot decide whether to accept either of the
proposed mutual beliefs, it will select a focus of information-sharing
by invoking {\bf Select-Focus-Info-Sharing} (Figure~\ref{focus_algo})
on {\em Better-Than(Logic,Algorithms)}. Since acceptance of the only
piece of evidence provided by the user results in acceptance of the
top-level proposed belief (step 4.2), the algorithm will be applied
recursively to the child belief. Since it in turn has no children,
{\em Teaches(Smith,Logic)} itself will be selected as the focus of
information-sharing.

\begin{figure}
\centerline{\epsfysize=3.8in\epsffile{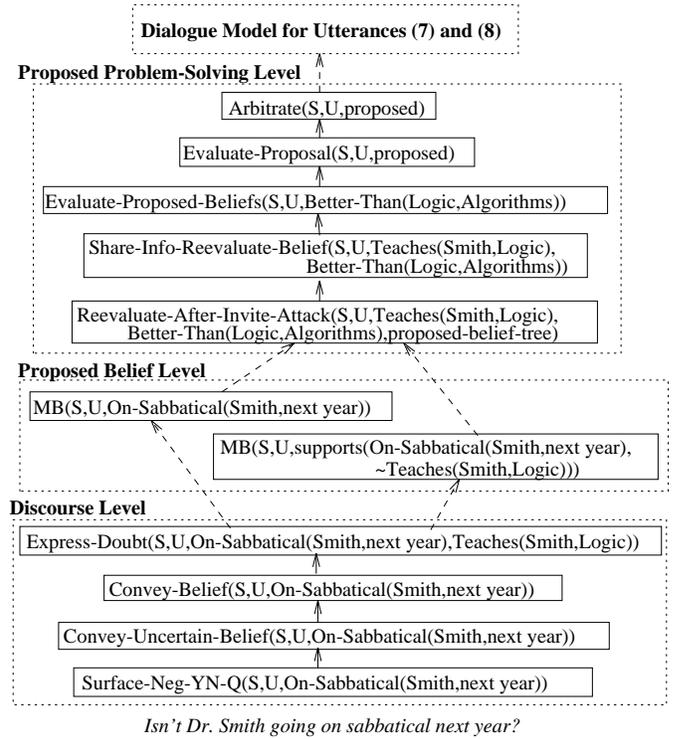}}
\caption{Dialogue Model for Generating Utterance
(\protect{\ref{sabbatical}})} 
\label{invite_attack}
\vspace{1ex}
\hrule
\end{figure}

The system will now invoke {\em Share-Info-Reevaluate-Belief} on the
identified focus.  Since the system's belief that Dr.~Smith is going
on sabbatical and its belief in the evidential relationship that being
on sabbatical implies that Dr.~Smith is not teaching Logic constitute
the only obstacle against its accepting {\em Teaches(Smith,Logic)},
they are considered a piece of {\em critical evidence}. Thus, {\em
Reevaluate-After-Invite-Attack} will be selected as the specialization
of {\em Share-Info-Reevaluate-Belief}. Figure~\ref{invite_attack}
shows the dialogue model that will be constructed for this process. In
order to satisfy the preconditions of {\em
Reevaluate-After-Invite-Attack} (Figure~\ref{recipes}), the system
will post {\em MB(S,U, On-Sabbatical(Smith, next year))} and {\em
MB(S,U, supports(On-Sabbatical(Smith, next year),
$\lnot$Teaches(Smith,Logic)))} as mutual beliefs to be achieved. The
system will adopt the {\em Express-Doubt} discourse action to convey
its strong belief in {\em On-Sabbatical(Smith, next year)} and its
warranted belief in {\em supports(On-Sabbatical(Smith, next year),
$\lnot$Teaches(Smith,Logic))}. However, the latter will not be
explicitly stated because the system believes that the user will
derive the evidential relationship based on stereotypical knowledge
and the structure of the discourse.\footnote{Walker
\cite{wal_coling94} has shown the importance of IRU'S (Informationally
Redundant Utterances) in efficient discourse. We leave including
appropriate IRU's for future work.} Thus the system would generate the
following utterance:

\bdialogat{S:}{}{8}
\em

\speakerlab Isn't Dr.~Smith going on sabbatical next year?

\edialog

\subsection{Possible Follow-Ups to Utterance
(\protect{\ref{sabbatical}})} 
\label{possible}

Consider the following alternative responses to (\ref{sabbatical}):

\bdialogcont{U:}{}
\em

\speakerlabsub \label{forgot} Oh, you're right. I forgot about that.

\speakerlabsubcont \label{notsab} He was planning on it, but he
told me that he had decided to postpone it until 1996.

\speakerlabsubcont \label{teach} Yes, but he got a good offer from the
department to teach Logic while he's on sabbatical. 

\speakerlabsubcont \label{postpone2} I thought he postponed his sabbatical
until 1996.

\edialog

\noindent Utterance (\ref{forgot}) will be captured as the user
accepting the proposal in Figure~\ref{invite_attack} and acknowledging
it. Utterances (\ref{notsab}) and (\ref{teach}) are both the result of
the user rejecting the system's proposed mutual beliefs and attempting
to modify them \cite{chu_car_acl95}. In (\ref{notsab}) the user
rejects {\em MB(S,U, On-Sabbatical(Smith,next year))}, while in
(\ref{teach}) the user rejects the evidential relationship {\em
MB(S,U, supports(On-Sabbatical(Smith, next year),
$\lnot$Teaches(Smith, Logic)))}. In the case where the user responds
with (\ref{notsab}), if the system accepts the proposed belief that
Dr.~Smith postponed his sabbatical until 1996, the system and the user
achieve the mutual belief {\em MB(S,U, $\lnot$On-Sabbatical(Smith,next
year))}. Thus, the precondition of the {\em
Reevaluate-After-Invite-Attack} action in Figure~\ref{invite_attack}
is satisfied, and the system would re-evaluate {\em
Better-Than(Logic,Algorithms)} taking into account the newly obtained
information. Notice that in this case the mutual belief achieved to
satisfy the preconditions of {\em Reevaluate-After-Invite-Attack} is
different from the ones the system attempted to achieve --- utterance
(\ref{sabbatical}) was generated as an attempt to achieve {\em MB(S,U,
On-Sabbatical(Smith, next year))}, but the result is that both the
system and the user accept {\em MB(S,U, $\lnot$On-Sabbatical(Smith,
next year))}. In this case, although the goal of the {\em
Express-Doubt} discourse action is not satisfied, the agents' mutual
belief achieves the higher-level goal that the {\em Express-Doubt}
action is intended to achieve, namely a precondition of {\em
Reevaluate-After-Invite-Attack}; thus the {\em Express-Doubt} action
is abandoned. This example shows how the precondition of {\em
Reevaluate-After-Invite-Attack} captures situations in which the user
presents counterevidence to the system's critical evidence and changes
the system's beliefs.

\subsection{Evaluating Utterance (\protect{\ref{postpone2}})}

\looseness=-1000
Utterance (\ref{postpone2}) will be interpreted as a case in which the
user is uncertain about whether to accept or reject the system's
proposal in (\ref{sabbatical}) and attempts to share information with
the system to re-evaluate the proposal. It proposes a mutual belief,
{\em Postpone-Sabbatical(Smith,1996)}, which will be evaluated by {\bf
Evaluate-Belief}. Suppose the system believes that Dr.~Smith is
spending next year at IBM, which is evidence against
Dr.~Smith postponing his sabbatical. Then the system cannot determine
the acceptance of the proposed belief, resulting in the need to
initiate an information-sharing subdialogue.  The focus of
information-sharing is {\em Postpone-Sabbatical(Smith,1996)} since it
is the only uncertain belief. The system will then select an
appropriate information-sharing strategy. Since the system does not
know the user's reasons for believing {\em
Postpone-Sabbatical(Smith,1996)}, but does have a piece of
non-critical evidence against the proposed belief, the third
information-sharing strategy will be selected. Thus the system would
query the user for support for the proposed belief and also provide
its evidence against the belief, leading to the generation of the
following utterances:

\bdialogcont{S:}{}
\em

\speakerlab Why do you think Dr.~Smith postponed his sabbatical
until 1996? 
\dialine Isn't he spending next year at IBM?

\edialog

\section{Related Work}

Grosz, Sidner and Lochbaum \cite{gro_sid_ic90,loc_acl91} developed a
SharedPlan approach to modelling collaborative discourse, and Sidner
\shortcite{sid_aaai94} formulated an artificial language for modeling
such discourse. Sidner viewed a collaborative planning process as
proposal/acceptance and proposal/rejection sequences. Her artificial
language treats an utterance such as {\em Why do X?} as a proposal for
the hearer to provide support for his proposal to do X. However,
Sidner's work is descriptive and does not provide a mechanism for
determining when and how such a proposal should be made nor how
responses should be formulated in information-sharing subdialogues.

Several researchers have studied the role of clarification dialogues
in disambiguating user plans \cite{vbetal_ci93,ras_zuk_cogsci93} and
in understanding referring expressions \cite{hee_hir_tr92}. Logan et
al. \shortcite{logetal_tr94} developed an automated librarian that
could revise its beliefs and intentions and could generate responses
as an attempt to revise the user's beliefs and intentions. Although
their system had rules for asking the user whether he holds a
particular belief and for telling the system's attitude toward a
belief, the emphasis of their work was on conflict resolution and plan
disambiguation. Thus they did not investigate a comprehensive strategy
for information-sharing during proposal evaluation. For example, they
did not identify situations in which information-sharing is necessary,
did not address how to select a focus of information-sharing when
there are multiple uncertain beliefs, did not consider requesting the
user's justifications for a belief, etc. In addition, they do not
provide an overall dialogue planner that takes into account discourse
structure and appropriately captures embedded subdialogues.

\section{Conclusion}

This paper has presented a computational strategy for collaborative
information-sharing in situations where the system's current knowledge
does not allow it to make a decision about whether to accept or reject
a user proposal. Our model includes algorithms for determining when
information-sharing subdialogues should be initiated and for selecting
a focus of information-sharing. The latter algorithm takes into
account both the effect of the acceptance of a piece of evidence on
the acceptance of the top-level belief, and the difficulty in
resolving the uncertainty about acceptance of a piece of
evidence. Furthermore, we have identified four alternative
information-sharing strategies and the criteria under which each
should be invoked, thus allowing the agents to share the most
pertinent information in order to re-evaluate a proposal. In addition,
by capturing information-sharing as part of the evaluation process in
a {\em Propose-Evaluate-Modify} cycle of actions, our model can handle
embedded information-sharing subdialogues.

\section*{Acknowledgments}

The research has benefitted from discussions with Stephanie Elzer,
Kathy McCoy, and Candy Sidner.

\bibliographystyle{named}

\end{document}